\pdfoutput=1
\documentclass[10pt,twocolumn]{article}

\usepackage{iccv}
\usepackage{times}
\usepackage{epsfig}
\usepackage{graphicx}
\usepackage{amsmath}
\usepackage{amssymb}
\usepackage{wrapfig}
\usepackage{array}
\usepackage{multirow}
\usepackage{tabularx,booktabs}

\usepackage[pagebackref=true, colorlinks,bookmarks=false]{hyperref}

\iccvfinalcopy 

\def\iccvPaperID{0023} 

\ificcvfinal\fi

\newcommand\ofirmodel{\textsc{OMG-Attack}}
\newcommand\fgsm{\textsc{FGSM}}
\newcommand\ifgsm{\textsc{I-FGSM}}
\newcommand\mifgsm{\textsc{MI-FGSM}}
\newcommand\vmifgsm{\textsc{VMI-FSGM}}
\newcommand\advgan{\textsc{AdvGAN}}
\newcommand\pgd{\textsc{PGD}}

\newcommand\mnist{\textsc{MNIST}}
\newcommand\traffic{\textsc{GTSRB}}
\newcommand\birds{\textsc{CUB-200}}

\newcommand\mnisttgt{MNIST-CNN}
\newcommand\birdstgt{CUB-ResNet18}
\newcommand\traffictgt{TS-CNN-STN}

\newcommand\mnisttra{MNIST-TR1}
\newcommand\mnisttrb{MNIST-TR2}
\newcommand\resnet{ResNet-18}
\newcommand\resnetf{ResNet-50}
\newcommand\wideresnet{WideResNet-50}

\begin{document}
\title{\ofirmodel: Self-Supervised On-Manifold Generation \\ of
Transferable Evasion Attacks}


\author{Ofir Bar Tal$\ \hspace{0.05cm}$~~~~~~~~~~~~~~~~~Adi Haviv~~~~~~~~~~~~~~~~~Amit H. Bermano\\ \\
The Blavatnik School of Computer Science \\
Tel Aviv University \\
\hspace{0.05cm}{\tt\small{ \{ofirbartal100,amit.bermano\}@gmail.com, adihaviv@cs.tau.ac.il}}
}

\maketitle
\ificcvfinal\thispagestyle{empty}\fi

\begin{abstract}

Evasion Attacks (EA) are used to test the robustness of trained neural networks by distorting input data to misguide the model into incorrect classifications. Creating these attacks is a challenging task, especially with the ever increasing complexity of models and datasets. In this work, we introduce a self-supervised, computationally economical method for generating adversarial examples, designed for the unseen black-box setting. Adapting techniques from representation learning, our method generates on-manifold EAs that are encouraged to resemble the data distribution. These attacks are comparable in effectiveness compared to the state-of-the-art when attacking the model trained on, but are significantly more effective when attacking unseen models, as the attacks are more related to the data rather than the model itself. Our experiments consistently demonstrate the method is effective across various models, unseen data categories, and even defended models, suggesting a significant role for on-manifold EAs when targeting unseen models.

\end{abstract}
\section{Introduction}
\label{sec:intro}

Evasion attacks are a fundamental tool for model robustness validation \cite{Croce2020RobustBenchAS, Croce2020ReliableEO, Shao2022OnTA}. As neural networks become increasingly ubiquitous, analyzing trained models for weaknesses by finding EA becomes more and more crucial. Despite their value, EAs are still disregarded during typical model development. This is because EAs are challenging to generate, especially as models become larger \cite{caron2021emerging,xie2017aggregated,he2022masked} and datasets more elaborate \cite{lin2014microsoft,van2018inaturalist,liu2015faceattributes}. 

\begin{figure}[t]
\centering
\includegraphics[width=0.45\textwidth, bb=0 0 640 480]{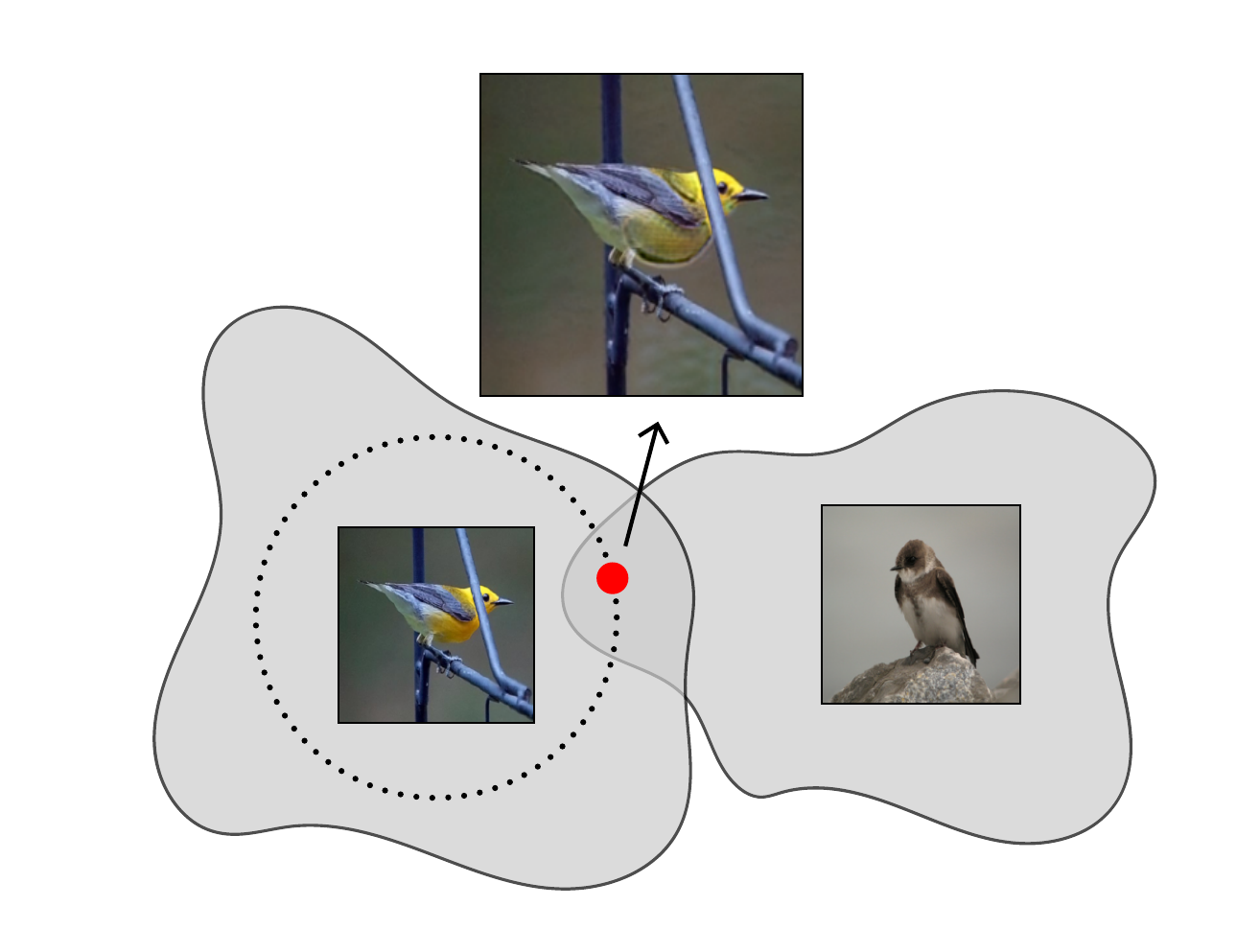}

\caption{Visualization of the on-manifold adversarial perturbation method on the \birds{} (Birds) dataset. The method generates examples on the general dataset manifold, but in an adversarial direction from the source image, projected on the $L_1$ ball, with a fixed radius \textit{budget}.}
\vspace{-18pt}
\label{fig:manifold-visualization}
\end{figure}

In terms of computational load, some approaches use optimization to generate individual attacks, which is infeasible for large-scale training \cite{Carlini2016TowardsET,moosavi2016deepfool}. Methods that produce EAs more efficiently are typically white-box based, where the attack generator leverages full knowledge and access to the target model, including its architecture, parameters, weights, and gradients \cite{Goodfellow2014ExplainingAH,madry2018towards}. Again, for real-life scenarios training an adversarial counterpart for each target model is infeasible. 

Black-box attacks are more suitable in this aspect, where the attacker has no knowledge of the target model parameters \cite{Carlini2016TowardsET,Szegedy2013IntriguingPO}. In their general variant, however, black-box attacks typically require access to the target model during inference. 

Evasion attacks' transferability is a trait that enables the same attack to be effective against other unseen target models, making black-box attacks practical in real-world applications. Despite it's clear merit, transferability is challenging; optimization-based attacks typically perform poorly on transferability \cite{Carlini2016TowardsET,Kurakin2016AdversarialML}. Gradient-based methods, on the other hand, generate more transferable adversarial examples \cite{Goodfellow2014ExplainingAH,mifgsm}, but they usually have low success rates making them less appealing to use in black-box scenarios \cite{Kurakin2016AdversarialML}.

In this paper, we present \ofirmodel{}, an EAs generation approach that addresses all aforementioned considerations. \ofirmodel{} is a self-supervised, computationally efficient, and \textit{transferable},
meaning it can be trained on one model, and used to attack another, alleviating the need to access the target model altogether.

To realize the method, our first insight is to turn to representation learning and data augmentations. In the representation learning setting, data points are embedded as spread away from each other as possible, regardless of labeling, while being agnostic to predetermined or learned augmentations \cite{bengio2013representation,shorten2019survey}. These augmentations can be considered evasion attacks generated to strengthen the representation during training. 

For example, the ViewMaker framework, an efficient, simple-to-implement, self-supervised generative approach for enhancing a representation learning process through automatic augmentation, was proposed by Tamkin \etal~\cite{tamkin2020viewmakernetworks}. In essence, the method predicts a perturbation of fixed strength that is most likely to change the representation of the input image. Using this framework, instead of learning a representation encoder, we learn augmentations that are designed to attack our pretrained target model. This results in a self-supervised, efficient, EAs generator.

\begin{figure}
\centering
\includegraphics[width=.45\textwidth,bb=0 0 520 303]{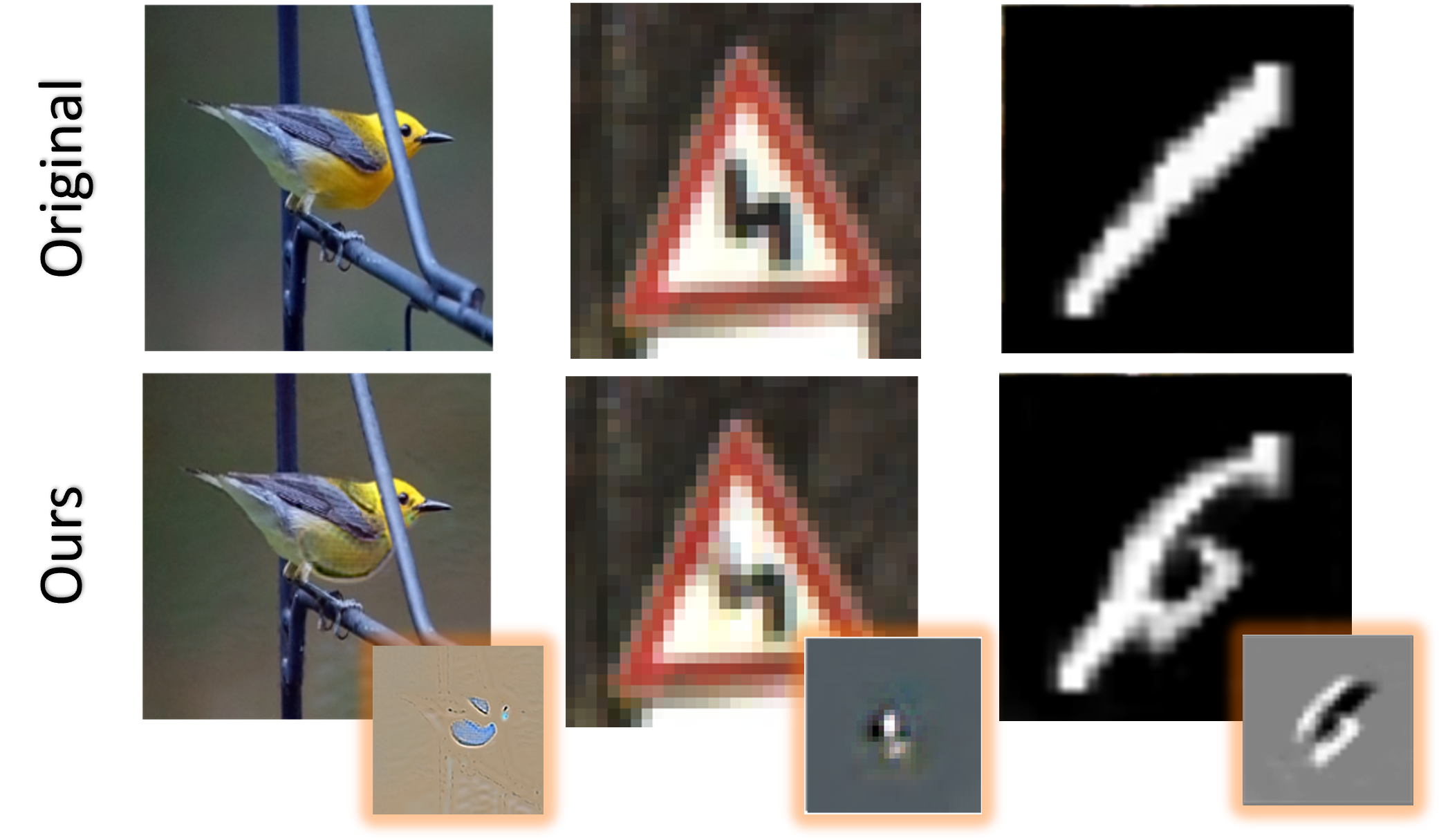}

\caption{Example of the \ofirmodel{} applied to each dataset. The upper row presents the original images, while the lower row shows the corresponding attacked images. The bottom right of each image exhibits the introduced perturbation.}\
\vspace{-10pt}
\label{fig:omg-visualization}
\end{figure}

Next, in order to promote transferability, our second insight is on-manifold generation. Intuitively, since on-manifold attacks are closer to the source distribution, they are attacks that are conceptually, affected more by the data distribution itself, rather than the target model and it's parameters. Hence, such attacks are effective for any model that is trained on the same data.  
In \ofirmodel{}, we encourage the EAs to remain on the data manifold, producing attacks that are closer to the data distribution and, in practice, seem more natural, as shown in Figure \ref{fig:omg-visualization}. 
We do this by adding a discriminator to the aforementioned ViewMaker framework \cite{tamkin2020viewmakernetworks}. This forces our attack generator to produce images that are indistinguishable from the source distribution in the eyes of the discriminator, hence producing on-manifold EAs. Indeed, we empirically show that on-manifold examples are transferable to unseen models, and even to new classes within the same dataset, that the \ofirmodel{} model has not seen during training. 
Through a series of experiments, we show that on-manifold attacks (Figure \ref{fig:manifold-visualization}) are highly transferable, demonstrating robustness to cross-model evaluation on several datasets. Compared to established baselines in off- and on-manifold EAs generation, we especially show promising attack success rates in the black-box settings on unseen defended and un-defended models. Additionally, we show superior performance for classes unseen during training for all of the evaluated data. 

Overall, these experiments show that on-manifold EAs are highly transferable and, in conjunction with a self-supervised training process, offer a viable solution for real-life use cases. 

\section{Related Work}
\label{sec:related}

\paragraph{Adversarial Examples} Early adversarial example generation techniques, such as L-BFGS \cite{Szegedy2013IntriguingPO}, \fgsm{} \cite{Goodfellow2014ExplainingAH}, and BIM \& ICML \cite{lee2017making} attacks, primarily focused on a \textit{white-box} context, where the attacker has full model insight including architecture, parameters, gradients, and training data. This led to the development of gradient-based methods, a current research focus.

Prominent white-box methods include FSGM \cite{Goodfellow2014ExplainingAH}, its iterative variants \cite{mifgsm, Wang2021EnhancingTT}, and \ifgsm{} by Madry \etal \cite{madry2018towards}, using the Projected Gradient Decent (\pgd{}) method. While most attacks constrained their $L_2$ or $L_\infty$ norms to maintain clean image visuals, others \cite{Papernot2015TheLO, Su2017OnePA} opted to generate perturbations via the network's forward gradient and saliency map.Our methodology aligns closely with the \textit{semi-white-box} or \textit{grey-box} attacks \cite{xiao2018generating, xu-etal-2021-grey, vivek2018gray,wang2021similarity}. In these methods, a generative model that has been trained (e.g., \cite{NIPS2014_5ca3e9b1}) is used to construct adversarial perturbations. This introduces an innovative approach where full access to the model is only required during training. During the inference phase, there's no need for access to the target model since the generator is already trained. In contrast to earlier works, our model leverages self-supervision in a realistic setting and ensures that the perturbations follow an approximated manifold.

\textit{Black-box} attacks present the most challenging scenario. These are further divided into score-based settings, where the attacker has access to the prediction distribution or confidence scores \cite{joshi2019semanticadversarial, Ma2020SimulatingUT}, and more stringent decision-based settings, where the attack model solely depends on the prediction's label \cite{Tolias2019TargetedMA,Brunner2018GuessingSB, Dolatabadi2020AdvFlowIB,Dolatabadi2020AdvFlowIB}. Query-based attacks are dominant in decision-based approaches. They estimate the gradient using zeroth-order optimization methods when the loss values can be queried, as proposed by Chen \etal \cite{chen2017zoozerothorder}. 

Although our work can be applied in a black-box setting during testing, it adheres to a stricter real-world setting where the attacker is decision-based and does not incorporate additional parameters or training steps, as in the subtitle transfer model approaches.

\paragraph{On-Manifold Adversarial Examples} The topic of generative adversarial examples has seen considerable interest \cite{chakraborty2021survey, akhtar2018threat, xu2020adversarial}. Of particular novelty is the generation of adversarial examples that adhere to the data manifold, a contrast to traditional adversarial attacks known to deviate from it \cite{song2018pixeldefend}. Gilmer \etal ~\cite{gilmer2018adversarial} provided initial evidence of the existence of on-manifold adversarial examples, and subsequent research has underscored the theoretical advantages of these examples in enhancing model robustness \cite{Patel2019OnmanifoldAD, Suggala2018RevisitingAR}. 

Goodfellow \etal ~\cite{goodfellow2020generative} suggested that optimally balancing the generator's predictive capabilities and the discriminator's enforcement of on-manifold constraints requires the generator to mimic the original data distribution. Inspired by this, Li \etal ~\cite{li2022review} proposed manifold projection methods using generative models. Stutz \etal \cite{stutz2019disentangling} introduced the concept of per-class on-manifold adversarial examples, though its application is limited in the absence of class labels. Several studies have leveraged generative models, such as Generative Adversarial Networks (GAN) \cite{NIPS2014_5ca3e9b1} and Variational Auto-Encoders (VAE) \cite{welling2014auto}, to generate on-manifold examples by introducing adversarial alterations to the embedding space \cite{Mangla2019AdvGANHL,patel2019onmanifoldadversarial,liu2020manigen}. Others, including AdvGan \cite{xiao2018generating} and AdvGAN++ \cite{Mangla2019AdvGANHL}, ensure perturbations remain on-manifold by imposing constraints, while some approaches rely on predefined semantic attributes \cite{joshi2019semanticadversarial}. 

Our method also uses a GAN-like structure akin to AdvGAN but differentiates itself through the application of the more modern viewmaker backbone, yielding better empirical results (\S\ref{sec:blackbox_results}). Finally, IDAA \cite{yang2022identitydisentangled} underscored the potential of on-manifold adversarial examples in achieving superior performance. In our research, we corroborate these findings through comprehensive evaluations, demonstrating that adherence to the data manifold can significantly boost the transferability of models to unseen models without necessitating additional training.

\paragraph{Self-supervised Adversarial Attacks}
Self-supervised learning, a methodology that involves training a model without predefined labels, has demonstrated significant potential across various computer vision tasks \cite{Gidaris2018UnsupervisedRL,Noroozi2016UnsupervisedLO, Chen2020ASF, He2019MomentumCF}. There are several paradigms that have been adopted in the realm of self-supervised learning. \textit{Instance Discrimination}, as exemplified in DINO \cite{caron2021emerging}, BYOL \cite{grill2020bootstrap}, and SimCLR \cite{krishnan2021eli5asimple}, makes use of common image processing augmentations to generate varied renditions of the same image. The objective is for models to create comparable representations or activations for these diverse versions. \textit{Masked Prediction} has found widespread use across fields such as Natural Language Processing \cite{devlin-etal-2019-bert,liu2019roberta}, Computer Vision \cite{Baevski2022data2vecAG}, and more. \textit{Transformation Prediction} is another paradigm centered around predicting image rotations \cite{Gidaris2018UnsupervisedRL}. Lastly, \textit{Clustering} methods have also been utilized in this space \cite{Caron2020UnsupervisedLO}.

The study of adversarial examples in representation learning has led to intriguing findings, with recent research indicating that the generation of adversarial examples can bolster the performance of self-supervised representation learning frameworks. The work demonstrated in CLAE \cite{https://doi.org/10.48550/arxiv.2010.12050}, for instance, established that generating adversarial examples could improve the representation learning process. Similarly, the authors of the ViewMaker network \cite{tamkin2020viewmakernetworks} proposed that adversarial examples could be effectively used as a substitute for predefined expert augmentations. 

In our study, we rely on and adapt the foundational model proposed in \cite{tamkin2020viewmakernetworks} to facilitate the generation of self-supervised adversarial attacks.

\section{The \ofirmodel{} Model}
\label{sec:method}

\begin{figure*}
  \centering
      \includegraphics[width=0.95\textwidth,bb=0 0 1081 349]{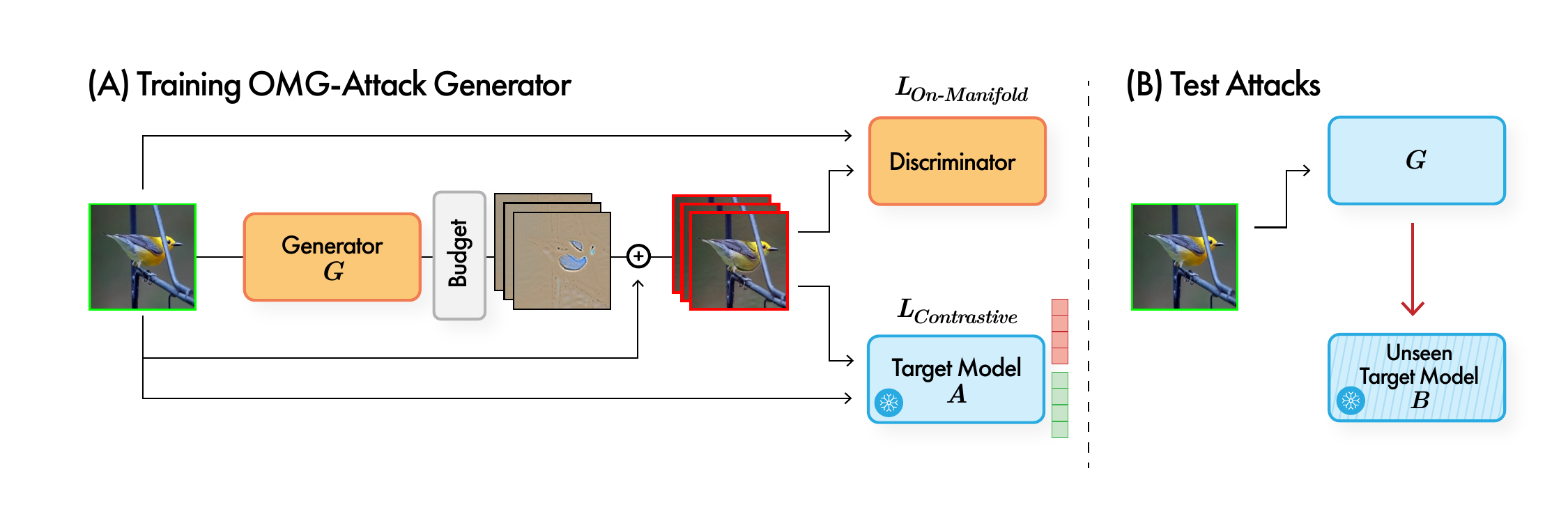}
  \caption{System Overview. (A) The Generator G creates adversarial perturbations of the input, projected onto an $L_1$ ball and scaled to adhere to a prescribed 'budget'. The Frozen Target Model A and the Discriminator process these images to train G. (B) the trained Generator G is used to perform adversarial attacks on the Unseen Target Model B.}



  \vspace{-10pt}
  \label{fig:pipeline}
\end{figure*} 

We now introduce our method of generating on-manifold non-targeted adversarial examples. 
In Figure \ref{fig:pipeline} we show our pipeline for learning and generating adversarial augmentations that are on-manifold examples.

\subsection{Problem Definition}

Let $X \subset I^{n \times n}$ be the image dataset of dimensions $n \times n$. Suppose that $(x_i, y_i)$ is the $i$-th instance within the training set, which is comprised of an image $x_i \in X$ , generated according to some unknown distribution $x_i \sim P_{data}$, and $y_i \in Y$ the corresponding true class labels. The learning system aims to learn a classifier $ f:X \rightarrow Y$ from the domain $X$ to the set of classification outputs $Y$, where $|Y|$ denotes the number of possible classification outputs. Given an instance $x \in X$, the goal of a regular adversary is to generate adversarial example $x_A$, which is misclassified $f(x_A) \neq y$ (i.e. untargeted attack),

In addition, the definition of a Manifold $M_y \subseteq I^{n \times n}$ is \[ M_y = \{x \in I^{n \times n} : C_y(x) = True \} \] where $C_y$ is the perfect binary classifier that decides if an image is of class $y$ or not. Thus $M := \bigcup_{y}M_y$ is the union of all classes manifolds in the dataset.

Therefore we further augment the regular adversary definition to include the on-manifold constraint. We aim to generate an adversarial example $x_A$, which is misclassified $f(x_A) \neq y$ and if part of the dataset manifold $x_A \in M$,

\subsection{Generator}
We use a dense prediction network as a generative model, inspired by the viewmaker networks \cite{tamkin2020viewmakernetworks}. The architecture of the model consists of an encoder-decoder structure, with residual blocks that incorporate random noise in the form of extra channels. 

Given an input image $x \in X$, the generator $G$ produces a perturbation $G(x,\eta)$ based on the input and a random noise vector $\eta \sim \mathcal{N}(0,1)$. The resulting perturbation is of the same dimensions and shape as the input image.





\paragraph{Applying a \textit{Budget} to Limit Perturbation Magnitude} An example is regarded as adversarial if it is perceptually indistinguishable from benign input but is classified differently (i.e. misclassified) \cite{Biggio2013EvasionAA,Szegedy2013IntriguingPO}. Typically by using $L_p$-norms to measure the perceptual similarity
between an adversarial input and its benign original \cite{Sharif2018OnTS,Papernot2015TheLO}. In our model, to scrutinize the region surrounding the input data point while simultaneously limiting the extent of modification, we introduce a fixed budget size for the perturbations. This budget size sets the limits for the perturbation magnitude, which is achieved by projecting the perturbations onto an $L_1$ ball and scaling it according to the pre-set budget size. The perturbation is subsequently defined as:

\begin{equation}
\centering
\delta = budget * \frac{\hat{G(X,\eta)}}{AvgMag(\hat{G(X,\eta))}} 
\label{eq:projection}
\end{equation}
where $AvgMag(v) = \frac{|v|_1}{n}$ and $|v|_1 = \sum_i^n{|v_i|}$. \\

Finally, the perturbation is superimposed onto the input image to create an adversarial example with augmentation, denoted as $X_{adv} = clamp(X + \delta,0,1)$. Here, $clamp$ ensures the resulting perturbation remains within the acceptable image domain. 
Typically the norm $L_\infty$ is employed  \cite{Kotyan2022AdversarialRA}, which means every pixel is individually limited in the amount it can change. However, this is prohibitive for on-manifold attacks, where the preferred mode concentrating the budget on strategic regions of the image. Indeed, in our experiments training did not converge when using $L_\infty$, as the discriminator was always able to identify the generated imagery. We therefore use the $L_1$ norm, as derived from the original view maker framework.

\subsection{Discriminator}

The discriminator, within our framework, is tasked with the crucial responsibility of learning the data manifold. Our discriminator's architecture is built upon the principles of the PatchGAN discriminator \cite{Isola2016ImagetoImageTW}, which primarily penalizes structural inconsistencies at the scale of local image patches. It aims to classify the authenticity of each patch in an image, categorizing them as either 'real' or 'fake'.

The PatchGAN discriminator operates convolutionally across the image, averaging all responses from individual patches to generate its final output. This approach essentially models the image as a Markov Random Field, assuming independence between pixels separated by more than a patch diameter.

Let $I$ be an input image and $D$ be our PatchGAN discriminator. The output from $D$ is a score matrix $D(I) \in \mathbf{R}^{n \times n}$, where each element $D(I)_{i,j}$ corresponds to the score attributed to a patch of size $p \times p$ centered at pixel $(i,j)$ in the input image $I$. The binary cross entropy loss (BCE) is utilized for each patch prediction:
\[BCE(p,y) = y \log(p) + (1-y) \log(1-p)\]
Penalize if a fake image is considered real, and vice versa (\textit{"real"} label is 1):
\[ l_{real} = BCE(D(x_{fake}),1)
, l_{fake}=BCE(D(x_{real}),0)\]
The final discriminator loss is the average of all patch losses:
\[ L_{Disc}=\frac{1}{2}(\frac{1}{n \times n}\sum{l_{real}} + \frac{1}{ n \times n}\sum{l_{fake}})\]


\subsection{Target Model}

Throughout the learning and training processes, our methodology retains full access to the target model. Notwithstanding, our distinctively crafted generator has the capacity to execute an attack during the inference stage, irrespective of the absence of the target model's loss. This aligns with the \textit{"semi-whitebox"} setting as detailed in \cite{xiao2018generating}. Opting not to utilize the logits from the target model, we leverage its penultimate representations instead. This strategic choice could feasibly extend the applicability of our framework to tasks beyond classification.



\subsection{Training via Self-Supervision}

Our pipeline uniquely operates on the learned representations rather than the classification layer, thereby obviating the need for labeled datasets. The training process for our pipeline is entirely self-supervised, leveraging two independent objectives to cultivate the generator's ability to manufacture adversarial, yet on-manifold, perturbations.

\paragraph{Contrastive Loss}
Instead of meddling with the predictions of the target model, our focus is on perturbing the learned representations. To accomplish this, we adopt a contrastive loss framework, a common technique in self-supervised and representation learning \cite{krishnan2021eli5asimple,caron2020unsupervisedlearning}. 

The objective here is to construct analogous representations for augmentations derived from the same source. Each input is associated with a positive example, while all other data points in the batch are regarded as negative examples. The loss is assembled based on the data points' similarity, with the optimal loss being obtained when the positive pair's similarity is 1 and the similarity with all other inputs is 0.


The contrastive loss, for $N$ groups of mutually exclusive positive pairs, is expressed as shown in the following:
\begin{align*}
\centering
L_{Pairs} &= \frac{1}{2N}\sum_{k=1}^{N}{l(2k-1,2k) + l(2k,2k-1)} \\
l(i,j) &=-\log{\frac{\exp{(s_{i,j}/\tau)}}{\sum_{k=1}^{2N}{1_{[k \neq i]}\exp{(s_{i,k}/\tau)}}}} 
\end{align*}

and $s_{a,b}$ is the cosine similarity of the representations of examples $a$ and $b$, and $\tau$ is a temperature parameter.

For all inputs in the batch, we use the generator to produce adversarial examples $X_{adv}^{i} = X^i+G(X^i,\eta)$. Examples generated from the same image are called positives, while all other examples in the batch (originals and generated) are considered to be negatives.

We compute this contrastive loss for three groups of positive pairs permutations: $(X,X_{adv_1})$, $(X,X_{adv_2})$, and $(X_{adv_1},X_{adv_2})$. This results in three losses:
\[ L_{Contrastive} = L_{Pairs}^{(X,X_{adv1})} + L_{Pairs}^{(X,X_{adv2})} + L_{Pairs}^{(X_{adv1},X_{adv2})} \]

Applying $L_{Pairs}$ between the two generated examples $(X_{adv1},X_{adv2})$ encourages the model to learn to produce diverse augmentations and avoid settling on a deterministic specific perturbation per input.

\paragraph{On-Manifold Loss}
Our approach for generating on-manifold, in-distribution augmentations involves incorporating an auxiliary loss function that leverages a discriminator model. Previous studies indicate that discriminators can efficiently learn the underlying manifold of a training dataset \cite{goodfellow2014generative,isola2017imagetoimage}. 


To integrate this into our methodology, we introduce an on-manifold loss as an adversarial objective for our generative network. The generative network aspires to generate augmentations that the discriminator deems part of the original dataset. As the discriminator is trained, it learns from its misclassifications and continually refines its comprehension of the dataset's underlying manifold.


Therefore, the on-manifold loss is derived from the discriminator's loss. It marks the areas where the generator needs to improve upon the fake patches detected by the discriminator: $L_{On-Manifold}=BCE(D(x_{fake}),0)$.

\paragraph{Total Loss}
The overall loss employed to train the generator is a combination of both the adversarial iteration of the contrastive loss and the on-manifold loss, which is the adversarial iteration of the discriminator objective:~$L_{total}~=~\alpha L_{On-Manifold}~-\beta~L_{Contrastive}$


Detailed information on the models' architectures and hyperparameters utilized in our study is provided in the supplementary material.
\section{Experiments} 

The goal of our experiments is to assess the efficacy of our model in generating adversarial examples. To that end, we evaluate our method by conducting experiments in various settings, including the white-box setting,  transferability to black-box models and architectures, including defended target models, and  transferability of attacks to unseen data classes at train time.

\subsection{Experimental Setup}
\paragraph{Datasets \& Models}
We evaluate our pipeline on three datasets from different domains, using target models of various scales and architectures for the classification task. The datasets used are presented in Table \ref{tab:datasets}.

\begin{table}[ht]
\small
\centering
\begin{tabularx}{\columnwidth}{XXXXX}
\toprule
\textbf{Dataset} & \textbf{Resolution} & \textbf{\#Classes} & \textbf{Data Type} & \textbf{Size}  \\ 
\midrule
MNIST & 28x28 & 10 & Digits& 60K \\ 
GTSRB & 32x32 & 43 & Traffic Signs &   40K \\  
CUB-200 & 224x224 & 200 & Birds & 12K \\ 
\bottomrule
\end{tabularx}
\setlength{\abovecaptionskip}{10pt}
\caption{An overview of the datasets used in our experiments. MNIST \cite{deng2012mnist} is a dataset of handwritten digits. GTSRB (German Traffic Sign Recognition Benchmark) \cite{stallkamp2012man} is a dataset consisting of images of traffic signs. CUB-200 \cite{He_2020} is the Caltech Birds dataset that includes images of 200 different bird species.}
\label{tab:datasets}
\end{table}

For each dataset, we train \ofirmodel{} (comprising a discriminator and generator) against a pre-trained model on the classification task of that particular dataset. For the \mnist{} dataset, we use the same target network architecture based on a Convolutional Neural Network (CNN) as used by Carlini \& Wagner \cite{Carlini2016TowardsET} (denoted as \mnisttgt{}), which has 1.2M parameters. For the German Traffic Sign Recognition Benchmark (\traffic{}), we use a CNN with three Spatial Transformers \cite{Garca2018DeepNN}, which achieves over 99\% accuracy on this dataset (denoted as \traffictgt{}) with 855K parameters. Lastly, for the Caltech Birds (\birds{}) dataset, we use the popular \resnet{} model \cite{He2015DeepRL} with 11.3M parameters (denoted as \birdstgt{}). 

As these models were used in training, we evaluate attacks on them under \textit{white-box} settings, where we have full access to the model parameters.

Under the \textit{black-box} settings, we use additional models to examine transferability. We leverage the \ofirmodel{} models trained under the white-box setting on each dataset separately and generate attacks for other unseen models. For MNIST, we use two models proposed by \cite{Tramr2017EnsembleAT} (denoted as \mnisttra{} and \mnisttrb{}) as well as a larger-scale \resnet{} trained on \mnist{}. For \traffic{}, we validate against a larger-scale \resnetf{}. Lastly, for the \birds{} dataset, we evaluate against a larger-scale \resnetf{} and \wideresnet{}.

Further details about the models, including the hyperparameters used, are provided in the supplementary material.

\paragraph{Baselines}
To comprehensively evaluate the effectiveness of our proposed model, we draw comparisons with an array of adversarial attack methods that span from simple, yet effective strategies to more complex and robust methods. This range of baselines allows us to benchmark our model against both traditional and state-of-the-art techniques.

\begin{enumerate}
    \item \textbf{Fast Gradient Sign Method (\fgsm)} \cite{Goodfellow2014ExplainingAH}: The method utilizes the gradients of the loss function with respect to the input data to create adversarial perturbations.
    
    \item \textbf{Projected Gradient Descent (\pgd)} \cite{Madry2017TowardsDL}: Building upon FGSM, PGD introduces iterative attacks to adversarial research. Unlike FGSM, PGD applies the perturbations multiple times within an epsilon ball, offering more potent adversarial examples with enhanced robustness.

    \item \textbf{Momentum Iterative Fast Gradient Sign Method (\mifgsm)}  \cite{Dong2017BoostingAA}: This advanced adversarial attack method infuses momentum into the iterative FGSM process. The momentum term allows the method to maintain the direction of gradient update across iterations, thereby enhancing both the effectiveness and robustness of the generated adversarial examples.

    \item \textbf{Variance-Tuning Momentum Iterative Fast Gradient Sign Method (\vmifgsm)}  \cite{Wang2021EnhancingTT}: As an extension to MI-FGSM, VMI-FGSM introduces a variance term to dynamically adjust the step size throughout the iterative process. This adaptive approach yields highly potent examples, thus placing VMI-FGSM among the most powerful adversarial attack methods to date.

    \item \textbf{\advgan} \cite{xiao2018generating}: A generative adversarial network-based method for creating adversarial examples. It employs a similar domain of operation as our proposed model, thus providing a pertinent comparison.
\end{enumerate}

\paragraph{Evaluation Metrics}
 We use the Attack Success Rate (ASR) to evaluate the effectiveness of our adversarial augmentations. ASR is defined as the fraction of adversarial examples that successfully cause the target model to change its prediction. Formally, 
 
 $$ASR = \frac{|\{X_{adv} | f(X) \neq f(X_{adv})\}|}{|\{X_{adv}\}|}$$
    
where $X_{adv}$ is an adversarial example generated from $X$. ASR is typically expressed as a percentage, with higher values indicating a more effective attack.

\begin{table*}[t]
\small
\centering
\begin{tabular}{llcccccc}
\toprule
Dataset & Model & \pgd{} & \fgsm{} & \mifgsm{} & \vmifgsm{} & \advgan{} & \ofirmodel{} \\
\midrule
\mnist{} & \mnisttgt{} & \textbf{99.87} & 87.48 & 99.86 & \textbf{99.87} & 89.27 & 99.37 \\ 
\traffic{} & \traffictgt{}  &  33.53 & 19.71 & 33.83 & 33.91 & 20.25 &  \textbf{67.48} \\
\birds{} & \birdstgt{}  & \textbf{94.27} & 92.85 & 93.65 & 93.89 & 75.09 & 46.97 \\
\bottomrule
\end{tabular}
\vspace{5pt}
\caption{White-box setting comparison. The Attack Success Rates (ASR) is reported, where all models are allocated a similar resource limit (i.e., budget). Without transferability, performance is comparable between models.}
\label{tab:basic-results}
\end{table*}


\begin{table*}[t]
\small
\centering
\begin{tabular}{llcccccc}
\toprule
Dataset & Def. Model & \pgd{} & \fgsm{} & \mifgsm{} & \vmifgsm{} & \advgan{} & \ofirmodel{} \\
\midrule
\mnist{} & \mnisttgt{} & 6.25 & 0.32 & 3.62 & 3.14 & 8.41 & \textbf{99.33} \\ 
\traffic{} & \traffictgt  & 16.75 & 8.71 & 17.41 & 17.65 & 6.99 & \textbf{64.58} \\ 
\birds{} & \birdstgt{}  & 12.65 & 14.58 & 19.35 & 25.43 & 4.75 & \textbf{38.94} \\ 
\bottomrule
\end{tabular}
\vspace{5pt}
\caption{Transferability to defended models. The Success Rates (ASR) is reported for attacks generated using undefended versions on models, but test on the models defended using adversarial training with FGSM. }
\label{tab:defended-models}
\end{table*}

\begin{table*}[t!]
\small
\centering
\begin{tabular}{llcccccc}
\toprule
Dataset & Model & \pgd{} & \fgsm{} & \mifgsm{} & \vmifgsm{} & \advgan{} & \ofirmodel{} \\
\midrule
\multirow{3}{*}{\mnist{}} & \mnisttra{} & 76.98 & 55.84 & 80.65 & 84.15 & 82.64 & \textbf{99.29} \\
& \mnisttrb{} & 60.59 & 85.06 & 88.12 & 90.27 & 72.35 & \textbf{99.56} \\ 
& \resnet{} & 70.91 & 56.43 & 72.49 & 73.60 & 72.32 & \textbf{89.65} \\
\cmidrule{1-8}
\traffic{} & \resnetf{} & 10.74 & 8.70 & 11.65 & 11.76 & 9.01 & \textbf{58.89} \\ 
\cmidrule{1-8}
\multirow{2}{*}{\birds{}} & \resnetf{} & 31.12 & 37.90 & 45.84 & \textbf{54.20} & 17.63 & 48.33 \\ 
& \wideresnet{} & 25.14 & 33.42 & 38.67 & 46.02 & 19.93 & \textbf{46.84} \\
\bottomrule
\end{tabular}
\vspace{5pt}
\caption{Transferability to different architectures and scales. The attacks, originally generated for the white-box settings, are tested for their transferability to more robust black-box target models. The presented results are Attack Success Rate (ASR) percentages.}
\label{tab:transferability}
\end{table*}



\paragraph{Budget Selection}
Assuring comparability of the budgets across different methods is challenging, given the disparate $L_p$ norms utilized by these methods. Thus, we conducted an examination where we studied the influence of augmentations under identical budgets across various methods. For each dataset, we used the same budget across all methods evaluated: 0.3 for the \mnist{} dataset, 0.015 for the \traffic{} dataset, and 0.025 for the \birds{} dataset.

For the selected budget values, we found that the difference in the intensity of the augmentations (i.e., $P = 1/(nn) * \sum(I(x,y))^2$ for each pixel $(x,y)$ in the image I) is relatively small, less than 10\% among the methods. We discuss this challenge further in \S\ref{sec:limitations}, and additional information on the budget analysis and selection process can be found in \S\ref{sec:ablations}.

\paragraph{Base-line White-box Setting}

We compare the ASR performance of the trained models in the white-box settings against the different baselines and present the results in Table \ref{tab:basic-results}. As can be seen, the performance of our method is comparable to other approaches in this setting.

\subsection{Transferabilty to Unseen Models }
\label{sec:blackbox_results}
These experiments have been conducted in a black-box setting, where the target model's parameters, logits, and architecture are inaccessible.

\paragraph{Defended Models} These subsequent experiments aimed to evaluate the robustness of the adversarial examples (originally generated for the undefended model) in manipulating the predictions of a defended model. The defensive strategy deployed in these experiments involved training the original model using a mixture of original data and adversarial examples produced by the \fgsm{} method on the static undefended model. Our approach outperformed other strategies in all the scenarios, demonstrating that the defensive method had a substantially reduced impact on our adversarial examples. The results are presented in Table~\ref{tab:defended-models}.
 
 \paragraph{Transferability to Architectures and Greater Scales} The black-box transfer experiment was designed to examine the efficacy of our generated adversarial examples when tested against unseen and more sophisticated models. The goal was to ascertain whether these adversarial examples specifically undermine the unique target model they were trained on or whether they target challenging, under-represented areas in the data manifold that pose inherent classification challenges for any model.
To evaluate this, we devised a simple methodology that applies our data augmentations and assesses the performance of different models, independently trained from the \ofirmodel{} model, on these augmentations. The results are tabulated in Table \ref{tab:transferability}, where we observe that our augmentations achieve an ASR higher than the baseline by 10\%-40\%. The high ASR values suggest that the tested models encounter difficulties in accurately classifying the augmented data and do not significantly outperform the original white-box target model. This suggests that the adversarial examples are generalizable to unseen models. Notably, all tested models were more sophisticated, possessing a larger parameter space, supporting our hypothesis that the adversarial examples are located in less represented regions of the dataset's manifold.

\begin{table}[t]
\centering
\small
\resizebox{\columnwidth}{!}{%
\begin{tabular}{lcccccc}
\toprule
& \multicolumn{3}{c}{\advgan{}} & \multicolumn{3}{c}{\ofirmodel{}} \\
Dataset & Train & Seen & Unseen & Train & Seen & Unseen \\
\midrule
\mnist{} & 81.03 & 72.20 & 31.69 & 85.56 & 84.25 & \textbf{46.49} \\
\traffic{} & 20.55 & 12.89 & 4.13 & 69.19 & 65.54 & \textbf{60.61} \\
\birds{} & 31.11 & 24.21 & 24.42 & 66.98 & 63.90 & \textbf{70.09} \\
\bottomrule
\end{tabular}
}
\setlength{\abovecaptionskip}{10pt}
\caption{Domain transfer from 80\% to 20\%. The 'Train' column represents the ASR percentage on the truncated dataset, 'Seen' shows the ASR percentage on the untruncated dataset but only on the subset with seen classes, and 'Unseen' is the ASR percentage on the subset with unseen classes.}
\label{tab:domain-transfer}
\end{table}

\begin{figure}[t]
  \centering
  \includegraphics[width=0.3\textwidth,bb=0 0 1814 1889]{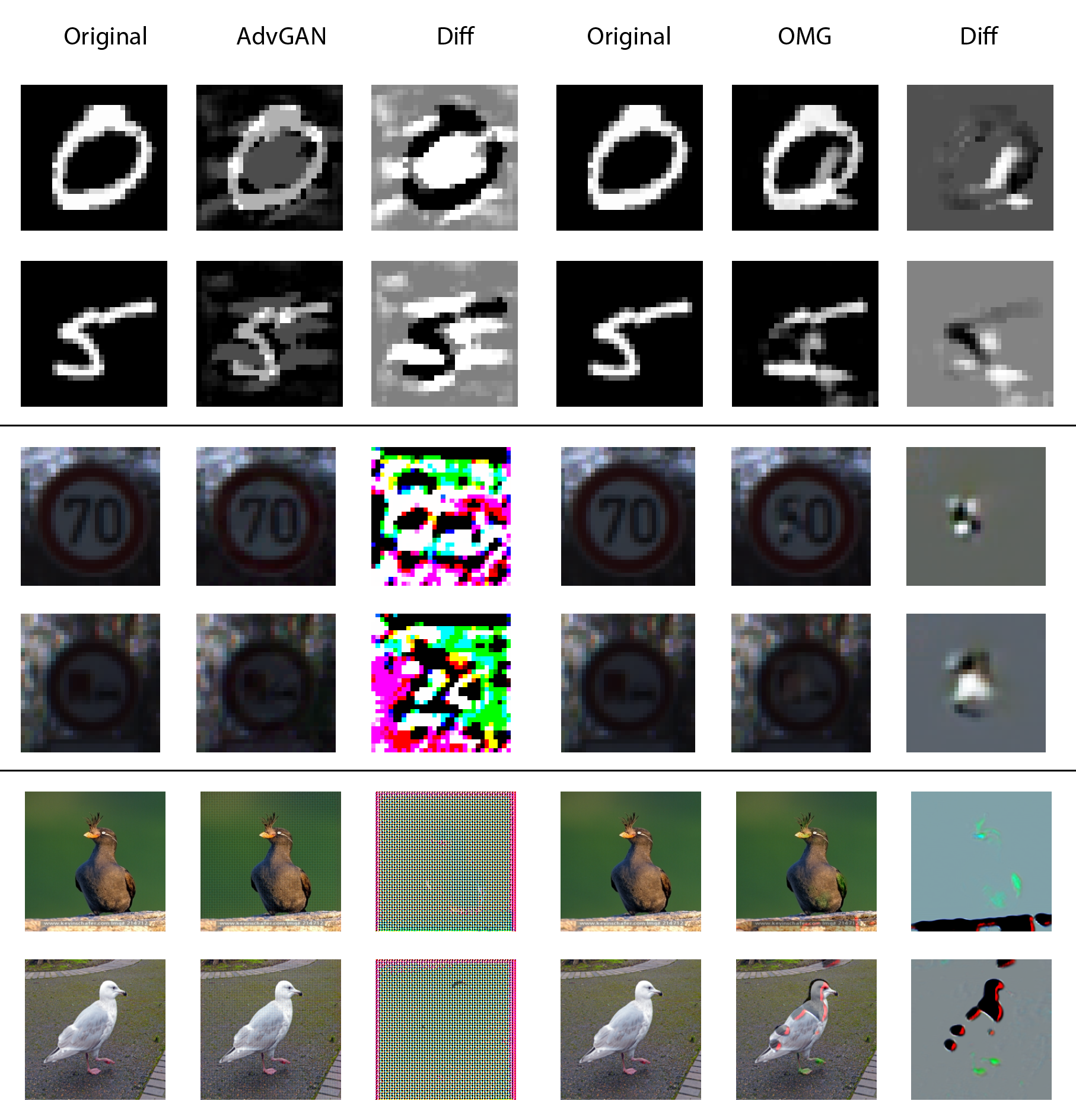}
  \caption{Input examples from classes that were not seen when the model was trained, show the ability of \ofirmodel{} to generalize to unseen data classes.}
  \label{fig:unseen_showcase}
\end{figure}


\subsection{Transfrability to Unseen Data} 
Another key advantage of our system is the ability to generate adversarial augmentations for classes that have not been seen by the target model. 
To evaluate this capability, we conduct an experiment where the target model is trained on only 80\% of the classes in a dataset. Subsequently, \ofirmodel{} is trained model, and utilized to generate augmentations for the entire dataset, including the previously unseen 20\%. We can see that our method consistently outperforms the baseline, by over 10\%-50\% in ASR. The results are reported in Table~\ref{tab:domain-transfer}, and depicted in Figure~\ref{fig:unseen_showcase}.

\section{Budget Analysis}
\label{sec:ablations}

The budget determines how much of the image can be altered, or how much of the signal must be preserved, and its magnitude greatly influences the model's performance. The higher the budget, more successful attacks (higher ASR) can be anticipated, but at the cost deviation from the original images and their content.

\setlength{\columnsep}{2pt}
\begin{wrapfigure}[13]{r}{0.25
\textwidth}
\centering
\includegraphics[width=\linewidth, bb= 0 0 342 318]
{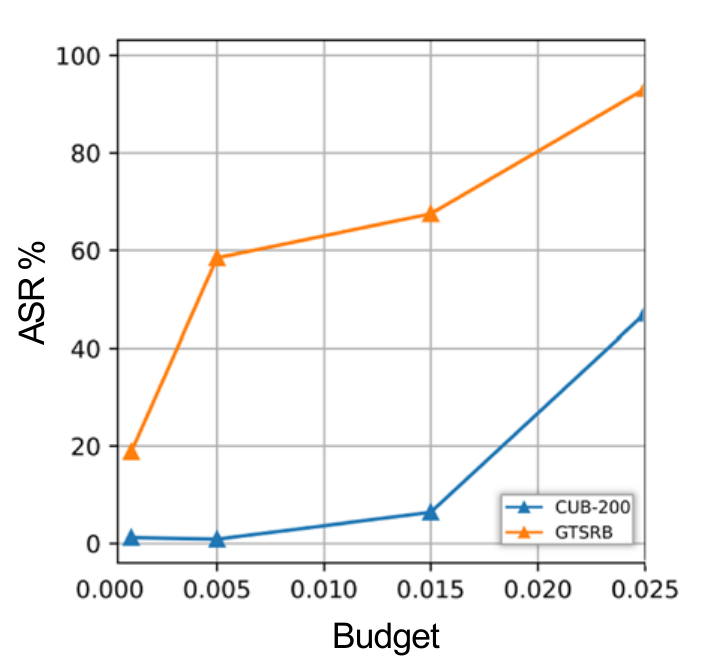}
\label{fig:budget}
\end{wrapfigure}
In the inset figure, we illustrate how varying the budget size influences the attack performance, particularly in terms of the ASR. 
Generally, a larger budget tends to produce adversarial examples that are more challenging for the model to classify correctly. 
This observation aligns with expectations since a larger budget provides the generator with greater flexibility to drastically alter the original image.

\section{Limitations}
\label{sec:limitations}

The primary limitation of our study revolves around the issue of budget selection. Our approach deviates from the majority of comparable baselines due to the unique normalization function we employed. While most standard baselines—including but not limited to AdvGAN—leverage the $L_{\infty}$ norm for budget allocation, our method uniquely utilizes the $L_1$ norm. Despite our concerted efforts to ensure similar budget allocations across methods for fair comparison, we acknowledge that the comparison might not be entirely balanced due to the disparate norms used.

A further limitation is that certain adversarial examples generated by our method, particularly within the \mnist{} dataset, lead to modifications that are noticeably perceptible to human observers. This challenges their classification as genuine adversarial examples, suggesting that a more rigorous budget constraint might be necessary for these types of datasets. This observation echoes the findings of previous research that has shown perceptible variations in \mnist{} dataset images, even when the $L_p$ norm of their differences is below thresholds utilized in earlier works \cite{Sharif2018OnTS}. The evidence, therefore, suggests that these changes can still significantly alter human perception of the images. 

In conclusion, while our method has shown promising results, these limitations highlight the need for further exploration in the area of budget allocation and the generation of truly imperceptible adversarial examples.

\section{Conclusions}
\label{sec:conclusion}

Adversarial augmentations serve as an insightful method for uncovering weaknesses in machine learning models. By directing adversarial perturbations to adhere to the data manifold, we can design data-dependent attacks that have transferability to various models. 

In this study, we proposed a self-supervised framework for learning and creating such attacks. Our approach draws upon adversarial training, guiding the generator to produce perturbations that challenge the encoder while also remaining within the bounds of the manifold.

Our experiments illustrate the potential of our method in generating transferable adversarial examples, providing a closer look at the visual semantics these augmentations carry. Our findings highlight the need to account for data-dependent attacks in developing and evaluating machine-learning models. Overall, our work adds to the ongoing discussions about adversarial attacks and their effects on machine-learning models. We hope our insights encourage further research in this field, leading toward machine learning models that are more resilient to adversarial perturbations.

{\small
\bibliographystyle{ieee_fullname}
\bibliography{egbib}
}

\end{document}